# The behavioural translation style space: Towards simulating the temporal dynamics of affect, behaviour, and cognition in human translation production

Michael Carl[*], Takanori Mizowaki[§], Aishvarya Raj[||], Masaru Yamada[§], Devi Sri Bandaru[*], Xinyue REN[#]

[*]Kent State University, [§]Rikkyo University, [||] Independent researcher, [#]City University of Hong Kong

***Abstract***
*The paper introduces a novel behavioural translation style space (BTSS) that describes possible behavioural translation patterns. The suggested BTSS is organized as a hierarchical structure that entails various embedded processing layers. We posit that observable translation behaviour – i.e. eye and finger movements – is fundamental when executing the physical act of translation but it is caused and shaped by higher-order cognitive processes and affective translation states. We analyse records of keystrokes and gaze data as indicators of the hidden mental processing structure and organize the behavioural patterns as a multi-layered embedded BTSS. We develop a perspective in which the BTSS serves as the basis for a computational translation agent to simulate the temporal dynamics of affect, behavioural routines and cognition during human translation production.*

## 1. Introduction

Cognitive sciences and translation studies have long suggested that translation involves multiple layers of processing that unfold simultaneously in the translator's mind. Dual-process theories, such as Kahneman's (2011) distinction between thinking fast and slow, that is, between System 1 (intuition), and System 2 (analytical) provide a foundational framework for this view. Evans & Stanovich (2013) extend this perspective, suggesting an interaction model of affective and cognitive processes in decision-making, while Ajzen's (1991) theory of planned behaviour links attitudes to behavioural intentions and actions, integrating subjective norms and perceived control. Together, these approaches highlight how intuitive, analytical, affective, and social-cognitive mechanisms may converge, producing a dynamic interplay of processes rather than a single, linear path.

Translation scholars have likewise (sometimes implicitly) suggested dual-process hypotheses and models of the translating mind (e.g. Gutt 2005; Tirkkonen-Condit 2005; Schaeffer & Carl 2013; Hubscher-Davidson 2017; Muñoz & Apfelthaler 2022; Robinson 2023). Several of these accounts assume that cognitive and affective processes to co-occur. For instance, relevance theory (Sperber & Wilson 1995; Gutt 2000, 2005) argues that affective cues help listeners infer contextual meaning, while proponents of embodied cognition (Hubscher-Davidson 2017; Rojo López & Caldwell-Harris 2023; Robinson 2023) suggest that emotional states are not merely external modifiers but integral to the mental processes underlying human translation. Building on this tradition, we posit three interrelated (ABC) dimensions of mental processes by which humans produce translations: Affective (feeling/emotion-related) states, behavioural (action/observation-related) routines, and cognitive (reflection/thought-related) processes.

Much of empirical translation process research (TPR) in the past decades takes for granted the idea that the temporal structure of the observed translation behaviour reveals the structure and interactions of internal, hidden mental processes that generate these data. We build on this hypothesis, investigating the empirical traces of these processes and their interactions, with the goal to build a computational agent that can simulate the observable human translation behaviour. A simulation approach may unveil which configurations of the ABC processes are suited to (re)produce the observed translation behaviours and to assess their explanatory power. A simulation allows us to manipulate those variable configurations to test or explore different processing strategies, which can help identify aspects of the human translation process that are crucial for achieving, for instance, fluency, accuracy, or contextual coherence. A simulation approach may not only validate theoretical models but also provides insights into the underlying dynamics of human cognition and emotion during translation.

Carl (2023, 2024, 2025) has proposed a novel framework to assess and evaluate ABC-related translation hypothesis as a simulation by means of a computational translation agent. A prototype version of this agent is implemented as a partially observable Markov decision process (POMDP, Heins 2022, Mizowaki et al. 2025). The hierarchically embedded layers of the agent's architecture unfold on different timelines. Lower-layer behavioural routines realize automatized behaviour (e.g. lexical choices), while higher-layer cognitive processes may unfold gradually over larger structures, paragraphs or sections. Predictive processing (PP, Seth 2021; Clark 2023; Hohwy 2016) and active inference (AIF, Friston et al. 2017; Parr et al. 2022; Pezzulo et al. 2024) posits that these embedded layers interact simultaneously through top-down predictions and bottom-up prediction errors, which would enable the agent to maintain coherence while being adaptively responsive to immediate sensory changes. For instance, affective (A) processes may generate predictions that guide the behaviour (B-layer), but when sensory (bottom-up) input clashes with predictions (top-down), cognitive/reflective thought (C-layer processes) may trigger a recalibration and re-integration of predictions and observations (Robinson 2023).

While we ultimately aim at developing, refining, and evaluating this multi-layered computational translation agent, in this article we introduce and develop a hierarchical space of embedded behavioural translation patterns that specify each of the ABC strata and their relation. We propose a behavioural translation style space (BTSS) that spans the set of all conceivable translation behaviours which an agent may realize. The BTSS consists of behavioural templates that specify the embedded structures of ABC processes. The (computational) agent will select and follow trajectories within this multi-layered BTSS and instantiate behavioural templates with gaze and keystroke simulations that reproduce the temporal dynamics of human translation processes.

The BTSS is an embedded multidimensional space of behavioural templates – capturing variations of the temporal structure in gaze-typing coordination – to formalize different possible translation styles. By navigating this hierarchical space, the computational agent would simulate a particular translation style impacted by the three dimensions of the ABC architecture. A simulation approach to TPR makes it possible to systematically test hypotheses regarding cognitive and affective dynamics underlying human translation processes that would otherwise be difficult to achieve. The BTSS will also allow us to relate, for instance, translator expertise, text complexity, or quality demands, with the temporal dynamics and interaction of affective, routinized, and reflective behaviour.

Section 2 provides a review of measures that have been suggested to assess ABC layers of the translating mind. Section 3 provides an overview over the various dimensions and layers of the BTSS. This section argues that a BTSS can be grounded in a few behavioural features, which can be aggregated in different ways, to constitute the various layers of the BTSS. Section 4 illustrates how a BTSS can be based on empirical data extracted from the CRITT TPR-DB[1] (Carl et al. 2016), while Section 5 puts the BTSS (back) into the context and usability of a computational translation agent.

## 2. Measures of translation processes

A plethora of metrics have been suggested to capture and quantify behavioural patterns of translators that are suited to discriminate between different levels of translation expertise, for texts with different levels of translation difficulty, different domains, expected translation quality, as well as individual working styles. Many of these measures relate to keystroke and gaze data.

### *2.1. Keystroke bursts and production units*

Dragsted (2005) investigates different typing behaviours in novice and experienced translators by analysing the length and frequency of pauses between successive keystrokes. She distinguishes between an analytic processing style, characterized by longer keystroke pauses and lower production speed, distinguishing it from an integrated processing style, marked by shorter pauses and higher production speed. Muñoz & Apfelthaler (2022) refine this approach by identifying three inter-keystroke intervals (IKIs) that differentiate automated translation routines, unintentional halts, and intentional stops. Building on their task segment framework, we introduce a distinction between keystroke bursts (KBs) and production units (PUs) in Section 3.1. and 4.3., allowing for a granular analysis of translation dynamics.

Jakobsen (2002) observes that students and professional translators exhibit distinct self-revision behaviours, where professionals engage in more strategic and efficient corrections. Alves & Vale (2011) build on this notion, introducing micro translation units and developing a more nuanced classification of editing procedures, which reveal systematic differences between novice and expert translators. Their findings suggest that, as translators gain experience, their editing strategies become more refined and cognitively streamlined. Carl (2021) applies the notion of micro translation unit to the word level, demonstrating that translation duration is influenced not only by the number and type of self-revisions but also by the intrinsic translation ambiguity of individual words. This research highlights how cognitive effort and linguistic complexity manifest in keystroke patterns, providing valuable insights into how translators navigate and resolve translation challenges.

[1]The CRITT Translation Process Research Database (TPR-DB) is a large repository of translation process data hosted by the Centre for Research and Innovation in Translation and Translation Technology (CRITT) that contains more than 5000 translation sessions with more than 600 hours text (mainly translation) production time of recorded keystroke data. The CRITT TPR-DB contains for each translation session several summary tables with numerous features that are suited to determine and quantify different translation styles on the various processing layers of the embedded translation architecture suggested in this paper. It can be downloaded free of charge following the instructions on the CRITT website https://sites.google.com/site/centretranslationinnovation/tpr-db.

### *2.2. Translation units*

The notion of coherent typing interrupted by a typing pause is fundamental in the conceptualization of translation units (Malmkjaer 1998; Tirkkonen-Condit 2005; Englund -Dimitrova 2005). Angelone (2010) posits a triadic translation process: problem recognition, solution proposal, and solution evaluation. In this model, a translation unit – i.e. a typing pause followed by a coherent sequence of keystrokes – would indicate how translators cycle through "Uncertainty management and problem solving bundles" (Angelone 2010: 18) whenever a difficulty arises, indicating metacognitive activity at the micro level of problem recognition and solution evaluation. Angelone's considerations of uncertainty management in translation can be complemented with findings from bilingualism which suggest that bilinguals activate all their languages automatically and non-selectively during language processing (Dijkstra & Van Heuven 2002). As a result, translators suppress unsuited alternatives, words or structures that are contextually inappropriate. This cognitive juggling—simultaneously activating multiple candidates and inhibiting those that do not fit—manifests in the ongoing cycle of translation generation, selection and evaluation, in particular when faced with translation difficulties.

This suggests—for our multi-layered ABC model of the translating mind—that translators engage in cycles of hierarchically embedded processes where the translation flow can be interrupted, for instance, by a local "problem nexus" (Angelone 2010:18) and trigger a sequence of "recognize–propose–evaluate" steps (Angelone 2010:18). For instance, at the affective layer a translator may recognize that a translation does not feel right, at the behavioural layer they may propose an alternative and amend the translation, and at the cognitive layer they can evaluate the revised outcome (Robinson 2023). Angelone also notes that professional translators tend to exhibit more consistent bundling of these triadic steps, whereas students often display more fragmented, unbundled cycles.

### *2.3. Gazing patterns*

The analysis of gaze patterns in TPR has mainly borrowed measures from psycholinguistics and reading studies. The usual fixation measures include:

- Fixation duration: Where longer time in a single fixation indicates processing difficulty.
- First-fixation duration: The time of the first fixation on a word or region.
- Gaze duration: The sum of all fixations on a word before moving to another.
- Total fixation duration: The cumulative time spent on a word (including regressions).
- Fixation count: The number of fixations on a word or region, reflecting processing demand.

These measures are frequently used in word recognition tasks, where shorter fixations are observed on frequent and predictable words, while longer fixations are produced on low-frequency or ambiguous words (Rayner 1998). In sentence processing research, longer fixations and regressions can be expected for syntactically complex or ambiguous sentences (Clifton et al. 2007; Schotter et al. 2012), while in translation research word and sentence processing is also impacted by translation difficulties. Since translators work concurrently with two texts in two windows, these classical measures are not ideal for translation, in particular since translators frequently switch between the source text (ST) and the target text (TT). Translators may also engage in concurrent ST or TT reading while typing, a behaviour that increases translation efficiency, i.e. shorter translation duration (Schaeffer et al. 2019). In Section 3.2., we suggest a novel method to classify gaze patterns as stretches of linear reading, re-fixation and scattered fixations. This novel classification, we believe, is well suited to characterize different gaze behaviours in translation.

### *2.4. Translation phases*

Jakobsen (2002) noticed that translation sessions can be fragmented into different translation phases: orientation, drafting, and revision. Based on these translation phases, Carl et al. (2011) and Dragsted & Carl (2013) suggest a taxonomy of different translation styles grounded in gazing behaviour. Several types of planners can be distinguished based on reading behaviour in the orientation phase (Jakobsen 2011)—before starting to draft the translations—while others read sentence-by-sentence. Dragsted & Carl distinguish between large-context planners who first read a sentence (or a long stretch of the ST) before starting the translation, while a head starter engages in translation production with only minimal look ahead. Others may still read ahead clause-by-clause before typing out their translations. However, a head starter would merely read an incomplete functional complex (i.e. a partial phrase) prior to translation production. As we will discuss in Section 4.4., this conception of translation phases substantially overlaps with our notion of translation states (Carl et al. 2024), and the notions of Hesitation, Orientation, and translation flow.

### *2.5. Translator styles*

Numerous authors assume individual translator styles[2] (Mossop 2007) as a source of variation in the behavioural data, e.g. pausing duration, number and type of revisions, gazing patterns, etc. Drawing on Jung's (1921) classification of personality types and the Myers-Briggs type indicator (Briggs 1962), Lehka-Paul (2020) explains different personal translation styles (that is, translat*or* styles) by their dominant psychological functions for information-processing and decision-making. A translator, she argues, will consistently engage in either Thinking—which may correspond to top-down C-layer processes—or Feeling, which hints at the dominance of bottom-up sensory-driven processes and a strong integration of the A- and B-layers. These individual preferences influence the number and length of pauses and the typing and pausing patterns during revisions, e.g. the number of deletions or substitutions, but see also Saldanha & O'Brien (2013: 147) for a critical assessment of the method.

In a (super) longitudinal study, Hansen (2013) observed no differences with respect to self-revision patterns when testing professional translators after ten years working in the translation market. Carl (2024, 2025) observes a large variation in the temporal structure of typing patterns between different translators and a high amount of consistency within each translator, indicating stability of individual translation style preferences.

[2]We make a distinction between a *translation style* as a particular trajectory through the BTSS and a *translator style* as set of personally/individually preferred behavioral translation features that subsumes possibly many specific translation styles.

To sum up, this body of keystroke and gaze-related TPR indicates that there seem to exist stable, personal translation styles (i.e. translator styles, cf. Saldanha & O'Brien 2013: 133) that can be determined by the:

- temporal (pausing/typing) structure of keystroke patterns;
- gazing patterns and eye-hand coordination (e.g. amount of concurrent reading and writing);
- self-revision behaviour, for instance during end revision;
- structure of translation phases (orientation, drafting, revision).

In the next sections we investigate how the different layers of processing strata could be processed in empirical behavioural data and integrated into a coherent multi-layered BTSS. We believe that the CRITT TPR-DB (Carl et al. 2016) is a suited framework for this endeavour. We thereby answer Alves & Albir (2025: 207) who contemplate that "[o]ne of the main problems in our discipline is the lack of previous experience along with the […] lack of standardized instruments": The CRITT TPR-DB precisely aims at overcoming this lack of previous experience and the BTSS may become part of a standardized instrument that builds on previous proposals and insights of processing strategies and processing units. On this background, the next sections investigate how these different layers of processing strata could be empirically populated and integrated into a coherent multi-layered BTSS.

## 3. Fragmenting behavioural translation data

The search for basic translation units has a long history in translation studies. Basic translation units have been sought in the translation product, i.e. relations between the source and the target texts, in the translation process, i.e. translation units that deal with the translator's activity, or in the relations between the two (Thunes 2017). Some scholars define process-based translation units as a "stretch of the source text that the translator keeps in mind at any one time, in order to produce translation equivalents in the text he or she is creating" (Malmkjaer 1998: 286). Other scholars operationalize process-based translation units in terms of behavioural data. We develop this latter view and argue that process data can be fragmented on several temporal layers.

### *3.1. Activity units, keystroke bursts and production units*

Carl & Kay (2011), for instance, decompose translation process data into units of gaze movements (fixation units) and units of writing, i.e. production units. Hvelplund (2016) defines behavioural attention units (AUs)

> as *uninterrupted processing activity* allocated either to the ST (ST gaze activity), the TT (TT gaze activity and/or typing activity) or to the ST while typing (ST gaze activity and concurrent typing). Transitions to and from an AU indicate shifts in processing activity, and the point in time at which the transition occurs is used to identify the end of one AU and the beginning of the next AU. (Hvelplund 2016: 156, our emphasis)

While Hvelplund's definition integrates gazing and typing activities in a seamless manner, it leaves unspecified how exactly an "uninterrupted processing activity" should be defined. In addition, several scholars wonder what "exact cognitive processes [are] taking place during the pauses" (Rosa et al. 2018: 18). Some authors suggest that different types of mental activities during the translation process may be associated with different pause values (Lacruz & Shreve 2014; Couto-Vale 2016). Indeed, numerous suggestions have been discussed in TPR to specify the lapse of time between successive inter-keystroke pauses – ranging from around 200ms to 5000ms (Kumpulainen 2005; Couto-Vale 2016) or even 10 seconds (Jakobsen 1999) – often with a view as to how automated/fluent translation production could be defined.

Some scholars have proposed a hierarchical pausing structure. According to Saldanha & O'Brien (2013:120), Bernardini (2001: 249) suggests a notion of "attention units [which] are better defined in hierarchical rather than sequential terms, with smaller units being processed within larger units". Previous issues of the CRITT TPR-DB assumed two static pausing values, on an AU level and a production unit (PU) level of 400ms and 1000ms, respectively (Carl & Kay 2011)[3]. In addition, some scholars propose a translator-relative threshold (e.g. Dragsted 2005) based on the translators' typing speed. Miljanovic et al. (2025) suggests the median within-word typing speed plus two STD as a translator-specific threshold to identify interruptions in the translation production flow. Muñoz & Apfelthaler (2020, 2022), suggest several temporally embedded layers of translation processes, two of which are translator-specific, based on the median typing speed (i.e. their inter-keystroke interval, IKIs) within words and between words. Here, we follow their definition to introduce a distinction between keystroke bursts (KBs) as shorter temporal units and production units (PUs) as typing units that consist of one or more KBs (and KBIs).[4] KBIs and PUBs interrupt and respectively break sequences of successive keystrokes into KBs and PUs as follows:

- A keystroke burst interruption (*KBI*) is defined as:

  $$\mathbf{2 * median}\ \boldsymbol{within{-}word\ IKI.}$$

  A KBI separates two successive keystroke bursts (KBs). Following Muñoz & Apfelthaler (2022: 26), we assume that KBIs (*respites* in their terminology) are "non-intentional" typing halts, in which "attention and/or resources being drawn away from typing". Similar to Miljanovic et al. (2025), in this study we define a within-word IKI as the pause between two successive alphanumerical keystrokes.[5]

- A production unit break (*PUB*) is defined as:

  $$\mathbf{3 * median}\ \boldsymbol{between{-}word\ IKI.}$$

  A PUB amounts to Muñoz & Apfelthaler's (task segment) pause, which is, according to them, an intentional typing break. Again, following Miljanovic et al. (2025), we define a between-word IKI as the pause between a non-alphanumerical keystroke and a successive alphanumerical keystroke.

[3]In the course of this work, we have adapted these thresholds so as to become consistent with KBIs and PUBs; see below.

[4]Muñoz & Apfelthaler use the terms *tasks* and *task segments* respectively. *Tasks,* in their terminology, also subsume internet searches and other forms of human-computer interaction. As we only deal with logs of keystrokes here, we prefer KBs and PUs, and respectively KBIs and PUBs, for the typing pauses that separate KBs and PUs. *Typing bursts* could be an alternative term for KBs, which is well-established in writing research. However, it has often been used for longer sequences of keystrokes separated by longer pauses of 1sec. or 2sec. We use keystroke bursts (KBs) to stress that the keystroke sequences are shorter, separated by relatively shorter pauses.

[5]In Carl (2024), we have computed these values, i.e. respites and (task segment) pauses, in a slightly different way. The method used here produces different but highly correlates values.

A KBI, in this definition, indicates the completion of learned sensorimotor pattern that requires little conscious effort (thus indicative of B-layer activities), while a PUB indicates an intervention of higher-level cognitive functions to account for error monitoring, reflection, adaptation to novel contexts, etc. (cf. Angelone 2010, above), thus indicative of the C-layer activities.

With a notion of "uninterrupted processing activity" (Hvelplund 2016: 156) that is delimited and defined by KBIs, we can adopt Hvelplund's (2016) definition of AUs (see above) to specify six types of activity units[6] (AUs, see Table 1): We distinguish between reading, writing, and concurrent activities and—as in previous work (Carl et al. 2016; Carl & Schaeffer 2017)—we add an additional AU of type 8 that accounts for breaks longer than a KBI in which no translation behaviour was observed.

Table 1: Types of AUs and colour code in Figures 1 and 2

| AU type | Coordination of reading & writing activity | AU colour in Figure 1 |
|---|---|---|
| 1 | ST reading | Blue |
| 2 | TT reading | Light green |
| 4 | TT production | Yellow (no occurrence) |
| 5 | ST reading with concurrent TT production | Red |
| 6 | TT reading with concurrent TT production | Dark green |
| 8 | No observed behavioural data longer than *KBI* | Black (no occurrence) |

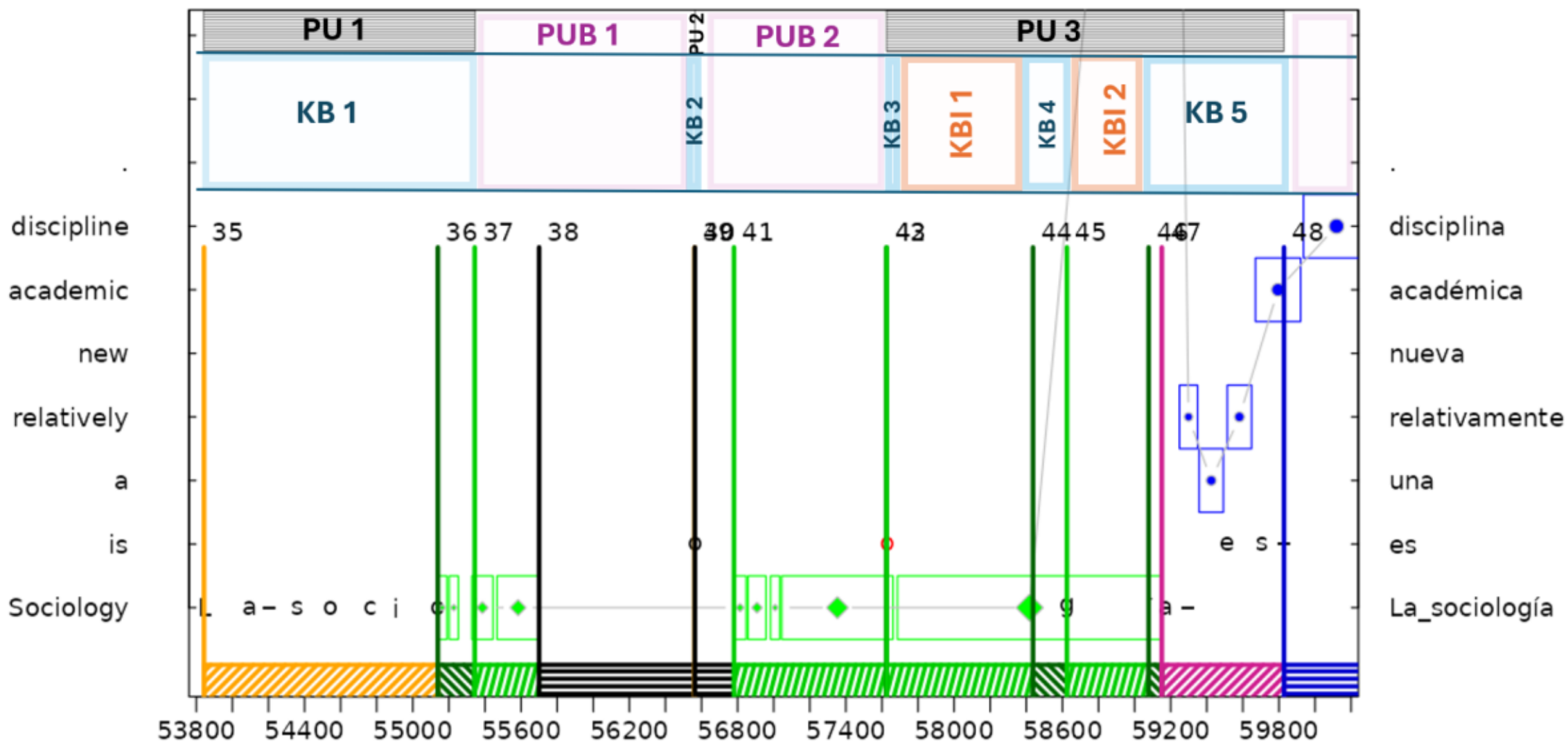

Figure 1: A progression graph of a small snippet of the translation session (BML12/P08 T5). The graph shows a segment of approximately 6 seconds (53,800–59,800 milliseconds) of an English-to-Spanish translation. The vertical axis plots to the ST on the left side and the TT on the right side, both in the ST word-order. ST and TT words are aligned on a word (or phrase level). A single ST word that maps into TT phrase is marked as a multi-word unit on the right vertical axis, such as *Sociology* ↔ *La sociología* (blank spaces are represented as underscores). Blue and green squares indicate eye

[6]While Hvelplund (2016) suggests the term *Attention Units* we prefer *Activity Unit,* as it better underpins the behavioral aspects of gazing and keystroke coordination, rather than an assumed cognitive process of selective focus and intention.

movements on the ST and TT respectively. The black and red characters (e.g. *o* around 57600) are insertion and deletion respectively. AUs fragment behavioural translation data into six categories (see Table 1), marked at the bottom of the graph in different colours and indicated by numbered, vertical lines (Id 35 to 48). Some AUs cluster into KBs (KB1 to KB5) marked at the top in the graph and intercepted with KB interruptions (KBIs) or PU breaks (PUBs). Sequences of KBs are grouped into production units plotted and numbered as grey boxes at the top of the graph (PU1 to PU3).

In order to illustrate the integration of translator-specific KBIs and PUBs with the notion of AUs and PUs, we discuss an example in Figure 1. The figure highlights the coordination of keystroke and gazing behaviour during the translation of "Sociology is" into Spanish "La sociologia es". Figure 1 shows a segment of approximately 6 seconds in which the part in bold of the English sentence "**Sociology is** a relatively new academic discipline." is translated into Spanish "**La sociología es** una disciplina académica relativamente nueva." The figure segments the gaze and keylogging data of this section into 13 AUs, numbered 35 to 47, as well as three PUs. While PUs are numbered PU1, PU2, and PU3 on top or the progression graph in grey striped bars, AUs are marked at the bottom in different colours (see explanation in Table 1) and marked by the numbered vertical lines. The computation of median within-word and between word IKIs provides translation-session specific KBI of 374ms and PUB of 891ms for this particular translation session.

Table 2 summarizes some of the main properties of the 13 AUs. Note that a KB can consist of several AUs. Thus, AUs 35 and 36 are two parts of one KB ("La_socilo"), one AU of Type 4 and a successive AU of Type 6. These two AUs contribute to one KB (only insertions [Ins]) and sum up to 1500ms, which is also the duration of PU1. PU1 is followed by a *PUB* of 1219ms, which is also distributed over two AUs (37 and 38) in which the translator's eyes first remain on the TT window (AU 37). No gaze data was collected during the following 862ms (hence AU 38 is of Type 8) in which the translator presumably looked at the keyboard. An *o* was then typed in AU 38, followed by a PUB of 1062ms. The PUB is also separated into two AUs of Type 8 and Type 2 with 217ms and 845ms respectively. The pressing of a single keystroke (keydown) (e.g. in AU 39) is a point in time for which we assume here a duration of 1ms. The *o* is again deleted in AU 42 (also duration of 1ms). The KBI of 811ms (AU43) separates deletion (AU42) from the following insertion of *og* in AU 44. Another KBI (453ms) follows, and a KB leads to the insertion of *ía es* in AUs 46 and 47. Three KBs and five AUs are aggregated as PU3.

The gaze in AUs 36 through 46 remains on the same TT word, whereas AU 47 shows how the translator reads until the end of the ST sentence. In AUs 47 and 48 the translator's eyes return to the ST, presumably taking in new information for successive translation while still typing the remainder of the chunk in AU 47. The translator needs to read ahead presumably, because the English sentence final word *discipline* swaps into an early position "una disciplina" in the Spanish translation.

AU sequences of Types 1, 2 or 8 cluster into one KBI if the total duration is larger than the KBI threshold, but smaller than PUB; they aggregate into one PUB if their total duration exceeds the PUB threshold. Successive AUs of Types 4, 5, or 6 (i.e. AUs which involve keyboard activities) are clustered into one KB. KBs may also include AUs of type 1, 2, or 8 if their durations are less than a KBI. The total gaze duration accumulated in an AU cannot exceed the duration of the AU. For instance, while the eyes remain on the TT in one extremely long fixation of 1500ms through four AUs (AU43 to AU46), this fixation is split into its respective parts of 786ms, 188ms, 453ms and 73ms and distributed over the four involved AUs (see Table 2).

Table 2: Properties of the 13 AUs (Id 35–47) depicted in Figure 1. The horizontal dotted lines separate three PUs and two PUBs. Each PU, PUB, KB, and KBI consists of one or more AUs. The Ids of the AUs are given in the Id column, while their types are provided in the AU column. AUs of Types1, 2, or 8 aggregate to pauses (PUBs or KBIs), AUs of Types 4, 5, and 6 aggregate into KBs and PUs. The right side of the table provides keystroke (number of insertions and deletions) and gaze data (number of fixations on the ST and TT (FixS and FixT) and the total reading time (TrtS and TrtT)), as well as the string produced (Edit) in the AU.

| Id | PU | KB | AU | Time | Dur | Ins | Del | FixS | TrtS | FixT | TrtT | Edit |
|---|---|---|---|---|---|---|---|---|---|---|---|---|
| 35 | 1 | Ins | 4 | 53843 | 1296 | 7 | 0 | 0 | 0 | 0 | 0 | La_soci |
| 36 | | | 6 | 55139 | 204 | 2 | 0 | 0 | 0 | 3 | 116 | ol |
| 37 | PUB 1 | | 2 | 55343 | 357 | 0 | 0 | 0 | 0 | 2 | 334 | --- |
| 38 | | | 8 | 55700 | 862 | 0 | 0 | 0 | 0 | 0 | 0 | --- |
| 39 | 2 | Ins | 4 | 56562 | 1 | 1 | 0 | 0 | 0 | 0 | 0 | o |
| 40 | PUB 2 | | 8 | 56563 | 217 | 0 | 0 | 0 | 0 | 0 | 0 | --- |
| 41 | | | 2 | 56780 | 845 | 0 | 0 | 0 | 0 | 4 | 796 | --- |
| 42 | 3 | Del | 6 | 57625 | 1 | 0 | 1 | 0 | 0 | 1 | 1 | [o] |
| 43 | | KBI | 2 | 57626 | 811 | 0 | 0 | 0 | 0 | 2 | 786 | --- |
| 44 | | Ins | 6 | 58437 | 188 | 2 | 0 | 0 | 0 | 1 | 188 | og |
| 45 | | KBI | 2 | 58625 | 453 | 0 | 0 | 0 | 0 | 1 | 453 | --- |
| 46 | | Ins | 6 | 59078 | 73 | 1 | 0 | 0 | 0 | 1 | 73 | í |
| 47 | | | 5 | 59151 | 677 | 5 | 0 | 5 | 684 | 0 | 0 | a_es_ |

Thus, Figure 1 and Table 2 illustrate a hierarchy of three embedded processing layers, in which AUs form the basic entities. Sequences of AUs are clustered into KBs (and *KBIs*) and sequences of KBs (and *KBIs*) are aggregated into PUs (and *PUBs*). Each of these three embedded layers evolves on a different timeline and inherits properties of the embedded layer (see Carl 2024, 2025).

### *3.2. A novel classification of gaze patterns*

In addition to the features provided in Table 2, the characteristics of eye movements in AUs (and by extension in the other layers, too) can be classified not only by the number and duration of the fixations, but also in terms of gaze patterns they form. Here, we make a distinction between linear reading, re-fixations, or scattered fixations (and no gaze data).

*Linear reading* refers to a gazing pattern in which the eyes move consecutively in the text's reading direction (for instance, left to right for English), proceeding almost in a straight line. It typically indicates normal, continuous reading behaviour, such as when a translator is reading the source text without interruption or when reading through the target text in one go to revise it. An example of linear reading is shown in AUs 47 and 48 in Figure 1, where the translator's eyes are detected on the ST, reading the remainder of the sentence, as discussed above.

*Re-fixations/re-reading* describes a pattern in which the gaze remains in nearly the same position or moves slightly backward, indicating repeated fixations in roughly the same area. This corresponds to re-fixation or small regressions. If it occurs briefly, it may reflect lower-level adjustments such as correcting the landing position of the gaze, whereas if it continues for a longer duration, it can suggest more deliberate and deeper cognitive processes, like contemplating word choices or monitoring translation production. An example is shown in AUs 36/37 and AUs 41 through 46 where numerous TT fixations are detected, on the same position, in which the translator presumably monitors the production of "La sociologia" (translation of "Sociology"), while struggling with some minor typos.

A *scattered fixation* pattern is characterized by sporadic/irregular eye movements that do not fit into either linear reading or re-fixation. The gaze may move up or down by more than 150 px, or it may jump horizontally across long distances, resulting in scattered eye movements without an obvious coherent flow. This pattern differs from normal reading behaviour and can include situations where the gaze roams while searching for necessary information, or when only specific parts of the text are checked intermittently rather than reading it in sequence.

### *3.3. Affective translation states, translation phases and translator styles*

In addition to the three layers discussed in the previous section (AU, KB, PU), we suggest three additional layers for fragmenting the translation process—and thus translation process data—on still broader layers. Carl et al. (2024) propose a HOF taxonomy[7] that characterizes three translation states:

- Hesitation state (H): A state of hesitation is triggered in a moment of surprise, characterized by re-starts, re-fixations and re-reading, revisions or text modifications.
- Orientation state (O): In a state of orientation, a translator aims at gathering new information, characterized by linear, forward-reading, mostly on the ST.
- Flow state (F): In a flow state, translation production is smooth and fluent, as reflected by no or very short keystroke pauses and with minimal reading-ahead.

While KBs and PUBs are indicative of B- and C-layer processes respectively, HOF states indicate A-layer processes. In our current conception, Hesitation and Flow states subsume complete PUs, that is, HOF states cannot fragment one PU into several parts, so that every PU is part of only one HOF state. The underlying assumption is that a PU reflects either Hesitation or Flow but cannot simultaneously be a mixture of both. According to Mirlohi et al. (2011: 253), Flow is a mental state of mind that is characterized by "intense focus, cognitive efficiency, a perceived skills-challenge balance, immediate feedback, merging of action and awareness, a sense of control, enjoyment, the opinion that time passes quickly, clearly defined task objectives and a lack of self-consciousness." Flow, they say, is a culturally universal phenomenon, independent from social class and age group, but may be experienced individually differently.

Hesitation and Flow states can consist of several PUs and/or PUBs. HOF states capture the degree of affective involvement within the embedded B- and C-layer processes and can be modelled as a mechanism of precision-weighting between these layers. In this way, HOF states not only regulate the balance of attention and effort across behavioural and cognitive processes but also integrate them into a unified control architecture, effectively subsuming the B- and C-layer processes within a broader affective–cognitive framework (see also Section 5.3).

[7] Since the original submission of this paper, we have added an additional revision state (R) to the HOF (now HORF) taxonomy that is characterized by linear, forward-reading, mostly on the TT.

In a previous study, two annotators annotated 1813 AUs with HOF state information (Carl et al. 2024). These annotations were successively used to train a machine learning algorithm (a Random Forest) to automatically annotate translation sessions from the CRITT TPR-DB. We use these annotations in Section 4[8].

A layer on a still broader timescale was proposed by Jakobsen (2002). As discussed in Section 2, Jakobsen (2002) distinguished three translation phases: an orientation phase[9], a drafting phase and a revision phase. In most cases, a translation session starts with an orientation phase, followed by the drafting phase. The beginning of the drafting phase is defined by the first keystroke, while the end of the drafting phase (and thus the beginning of the revision phase) can be defined as point in time when the translation production of the last ST word is finished. In our currently proposed BTSS we integrate translation phases as an extra processing layer, although there are large overlaps with HOF states, as we show in Section 4.4.

Finally, a sixth layer in the BTSS is translator styles. Mizowaki et al. (2024) suggest differentiating between five translator styles. They used n=5 different features (keystroke insertions, deletions, AU duration, KBI, and PUB) to classify a set of 522 translation sessions into five different translator styles. For each of the five features and each translation session they computed a relative inverse coefficient of variation (RICV) as the ratio of a translator's ICV (mean/std) divided by the overall ICV. Their classification reveals that translators' translation styles can be classified as 0) rapid 1) deliberate 2) confident 3) balanced and 4) cautious translators. We discuss these styles in more detail in Section 4.5. We thus distinguish six embedded layers of the BTSS as summarized in Table 3.

Table 3: Summary of embedded BTSS layers

| **Layer** | **Focus** | **Function** | **Example** |
|---|---|---|---|
| AU<br>(activity units) | Sequences of keystrokes and/or fixations | Sensorimotor integration, eye-hand coordination | Monitoring typing progress in the TT window |
| KB<br>(keystroke bursts) | Groups of keystrokes | Fluent typing, automated / routinized processes | Typing "translation" without pausing |
| PU<br>(production units) | Chunks of cognitive processing | Thoughtful/reflective processing segments | Amending a translation after verifying meaning |
| HOF states<br>(Hesitation, Orientation, flow) | Affective aspects of foraging / pragmatic affordances | Maintain a flow state, mediation of emotion, problem-solving | Information input, smooth typing, problem-solving |
| Translation phases | Macro-level patterns on a document level | Structured translation workflow | Head starter vs. context planner |
| Translator styles | Individual preferences | Strategic differences in translation approach | Fast vs. cautious vs. experimental translator |

[8]In the meantime, since the initial submission of this article, we have implemented a rule-based approach to automatically annotate HORF states (HOF + Revision, see Footnote 7). The rule-based approach gives us better control over the properties of these HORF states and it allowed us to add the revision state. We have successively clustered sequences of HORF states into translation policies, an additional processing layer as suggested and discussed in Carl et al. (2024), which is, however, omitted in this article.

[9]Note that the term *Orientation* is used as the initial translation phase, and it is also used as a state in the HOF taxonomy. Even though the two meanings of *Orientation* are somewhat related, they unfold on different strata in the BTSS.

## 4. Building the behavioural translation style space

In this section we develop the BTSS based on behavioural data extracted from the CRITT TPR-DB. We briefly describe a set of basic features that span the BTSS and then illustrate properties of the six BTSS layers and some of their relations.

Table 4: Features of the behavioural translation style space (BTSS): overall AU duration (Dur), keystroke related features (Ins, Del, TGnbr) characterize AUs of Types 4, 5, 6 while gaze related features (Dur_L, Dur_R, Dur_S) characterize AUs of Types 1, 2, 5, and 6.

| Feature | Meaning | Interpretation |
|---|---|---|
| **Dur** | Duration in ms of AU | High: more coherent translation style |
| **Ins** | Number of insertion keystrokes | High: fluent typing |
| **Del** | Number of deletion keystroke | High: error correction/revision |
| **TGnbr** | Number of target words produced | High: large chunk translation |
| **Dur_L** | Duration of linear reading pattern | High: take-in of new information |
| **Dur_R** | Duration of re-fixation pattern | High: monitoring / reflection |
| **Dur_S** | Duration of scattered fixations | High: visual search / distraction |

### *4.1. The empirical data*

We follow an empirical approach to estimate the parameters of the BTSS and the relations between the six layers of the BTSS, making use of the CRITT TPR-DB (Carl et al. 2016, see Footnote 1). In its current conception, AUs form the basis of the BTSS, while the remaining five layers are composed of AU sequences. To build the base BTSS we extracted from-scratch translation sessions from the CRITT TPR-DB. We excluded sessions in which PUB values > 6000ms or KBI values > 2000ms, as these are most likely due to flaws in recording (e.g, translators make many excessive pauses). We also excluded translations into Japanese and Chinese, since there is currently no easy way to distinguish and compute within-word and between word IKIs.[10] This left us with 521 translation sessions from English into various languages, as described in Figure 2 in Section 4.2. These sessions consist of a total of 357,639 AUs that were automatically annotated with HOF labels using Random Forest classifier (Carl et al. 2024, but see also Footnote 8) and successively classified with a non-supervised classification method (K-means) into five types of translation styles, following the procedure documented in Mizowaki et al. (2024). A multivariate base-BTSS was then established with seven AU features shown in Table 4.

[10]Some languages, such as Japanese and Chinese, do not use white spaces to separate words. In addition, due to the large character set, there are special input methods (IMEs) that capture the keystrokes—usually a phonetic transcription of the words—which are then converted into the proper Japanese or Chinese characters. There is thus an additional keystroke-to-character-to-word mapping which may distort temporal relations and make it difficult to distinguish between within- and between-word keystrokes. Additionally, IMEs can also learn and adapt to typists.

### *4.2. Features of the BTSS*

The features Ins, Del, TGnbr carry keystroke-related information that may occur in AUs of Types 4, 5, or 6. Features Dur_L, Dur_R, and Dur_S are gaze-related and may occur in AUs of Types 1, 2, 5, or 6, depending on whether the gaze is detected on the ST or on the TT, and whether concurrent keyboard activities are observed. We use these features in log transformed versions or also as relative duration, see Table 5.

Table 5: Basic BTSS features: percentages of occurrences (Occur) per Type of AU, duration of AU (in ms) and proportion (in %) of linear reading (RelDur_L), re-fixation (RelDur_R), scattered fixations (RelDur_S), as well as average number of insertions (Ins) and deletions (Del) and the number of TT words produced (TGnbr)

| AU Type | Occur% | Dur | RelDur_L | RelDur_R | RelDur_S | Ins | Del | TGnbr |
|---|---|---|---|---|---|---|---|---|
| **1** | 13 | 1570 | 30 | 21 | 40 | 0 | 0 | 0 |
| **2** | 21 | 977 | 27 | 32 | 32 | 0 | 0 | 0 |
| **4** | 21 | 629 | 0 | 0 | 0 | 4.4 | 0.64 | 1.49 |
| **5** | 6 | 393 | 25 | 21 | 37 | 2.72 | 0.13 | 1.23 |
| **6** | 14 | 568 | 28 | 32 | 24 | 3.32 | 0.85 | 1.37 |
| **8** | 25 | 1584 | 0 | 0 | 0 | 0 | 0 | 0 |

Table 5 provides an overview over the seven AU dimensions that are the basic BTSS. The column Occur provides the percentage of occurrences of AUs in our dataset. AUs of Type 8 (no data recorded) are most frequent (25% of all AUs). One reason for this may be that several studies in the dataset are recorded without eyetracker. Such translation sessions just consist of Type 4 AUs (typing observed) and Type 8 AUs (no typing observed). That is, an AU Type 8 is a stretch of time longer than a KBI in which the translator is not typing. Somewhat surprisingly, however, is that Type 2 AUs are more frequent than Type 1 AUs (21% and 13% of the AUs respectively), indicating that translators seem to be more often engaged in TT reading than in ST reading. However, the average duration (Dur) of Type 1 AUs is almost double that of Type 2 AUs, which reveals that our translators spend approximately the same amount of time on ST reading as compared to TT reading, albeit the latter in a more interrupted manner.

With 125ms, 138ms, and 136ms per keystroke (Ins+Del), the typing speed is approximately similar across the three typing-related AUs (Types 4, 5, and 6). However, the average duration of these types of AUs—and thus the number of keystrokes per AU—is significantly different. AUs of Type 5 seem to be the shortest (393ms) and least frequent (6%). Type-6 AUs account for almost twice that amount (14%), while Type-4 AUs are the most frequent with 21%, probably due to the fact that some sessions don’t have gaze data at all. Type-4 AUs have also the longest average duration (Dur: 629ms), with on average the most insertions (Ins: 4.4). There also seem to be far more deletions (Del) in Type-4 and Type-6 AUs as compared to Type-5 AUs, indicating that monitoring typing activities (Type 6) occur more frequently when TT corrections are required, thus leading to more deletions. Type-5 AUs (concurrent ST reading and typing), in contrast, show the least number of deletions, on average one deletion every 3 seconds, as compared to 667ms per deletion for Type-6 AUs.

Also, there is on average a larger number of words produced (TGnbr) per ms in the Type-5 production mode as compared to the other types of AUs. With respect to gazing patterns, Table 5 reveals that Type-1 AUs have a slightly larger percentage of linear reading (RelDur_L), as compared to Type-2 AUs (30% vs. 27% respectively), and Type-2 AUs show more re-fixations (RelDur_R) than Type-1 AUs (32% vs. 21%). Type-1 AUs also show the largest proportion of scattered fixations (RelDur_S: 40%).

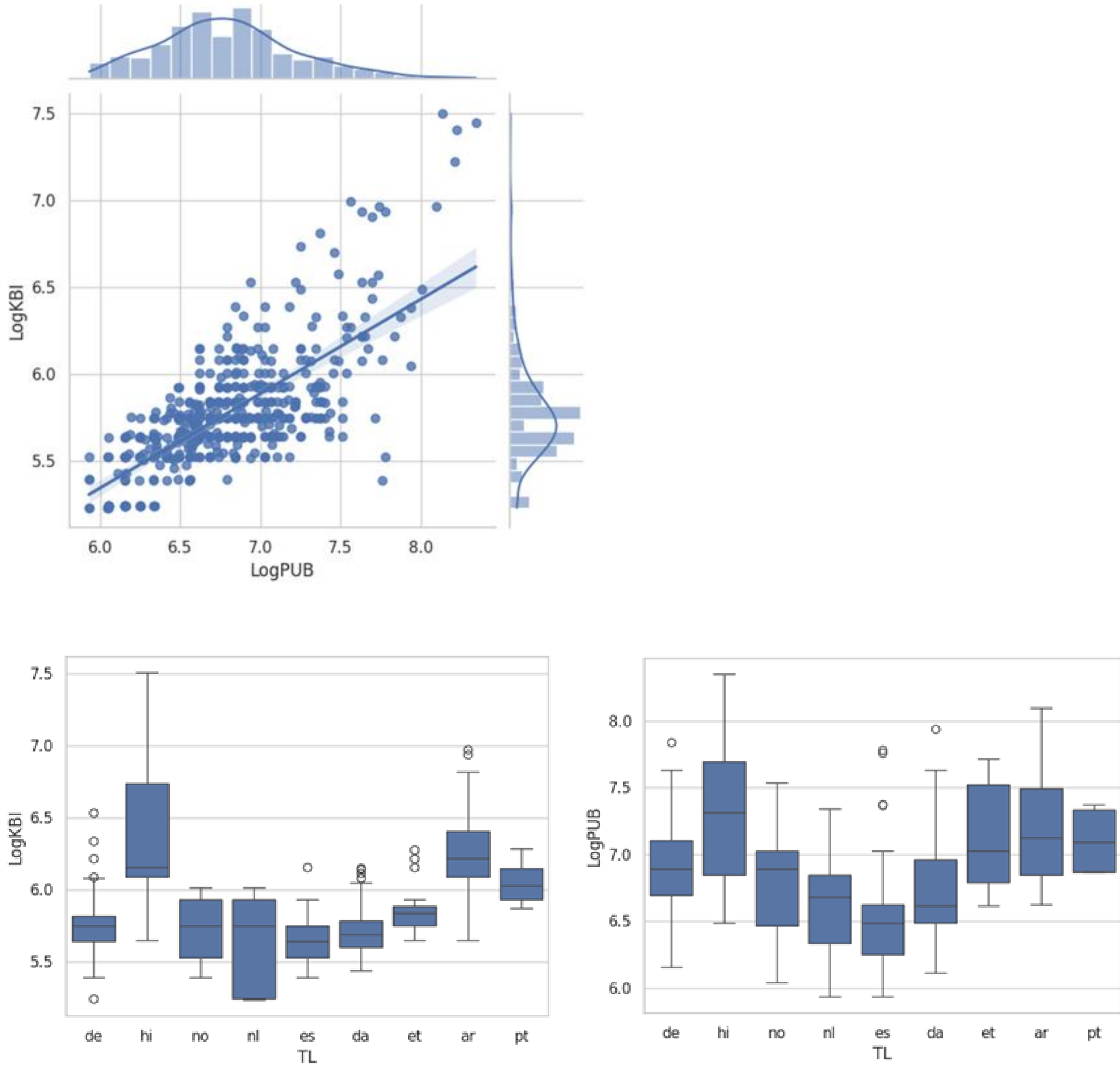

Figure 2: Relation between PUBs and KBIs across nine different languages. The scatterplot on the top shows the correlation for all 521 translation sessions, the boxplots break down the distribution of log KBIs and PUBs durations into the nine different target languages. Figure 5 provides a more fine-grained sub-classification of the Spanish LogPUB data (es) for personal translation styles.

### *4.3. KB interruptions and PU breaks*

Figure 2 plots the relation between KBIs and PUBs. As reported in several studies (e.g. Jakobsen 1999; Carl 2024, 2025), the rhythm of keystroke production in translation is largely language and translator dependent. While Carl (2024, 2025) investigates properties of the English to Arabic and Spanish translations, the two boxplots in Figure 2 show the distribution of KB interruptions and PU breaks for translations from English into seven European languages (German, Norwegian, Dutch, Spanish, Danish, Estonian and (Brazilian) Portuguese) and two non-European languages (Hindi and Arabic). As the plots show, there is clearly a difference in the typing (and thus pausing) speed between these different languages. Translations into Hindi are the slowest and most interrupted, while Spanish translators are the fastest in our dataset.

The scatterplot on the upper part in Figure 2 shows a strong correlation between PUBs and KBIs (r=0.72), indicating a high consistency of typing speed within words and between words (i.e. within and between linguistic boundaries) and the (assumed) temporal structure of unintentional and intentional typing halts. While it can be expected that some of these differences are due to different scripts (i.e. Arabic and Hindi), varying translator experiences and/or text difficulty among other reasons, we cannot dig into an analysis of the underlying reasons here. We just point out that there seems to be a systematic variation in the pausing/typing structure during translation production. Table 6 summarizes the variation in pausing across the 521 translation sessions. The table shows the minimum, maximum, mean, and median KBI and PUB values in ms and log ms. Note that sessions are excluded for which PUBs exceed 6 seconds and KBIs 2 seconds.

Table 6: Summary statistics for KB interruptions and PU breaks across 521 translation sessions from English into nine languages. KBI and PUB values are given in msec and LogKBI and LogPUB are in log msec.

| | LogKBI | LogPUB | KBI (ms) | PUB (ms) |
|---|---|---|---|---|
| Mean | 5.78 | 6.80 | 323 | 897 |
| Min | 5.23 | 5.93 | 186 | 376 |
| Median | 5.75 | 6.79 | 314 | 888 |
| Max | 7.50 | 8.35 | 1808 | 4230 |
| STD | 0.38 | 0.47 | 267 | 760 |

### *4.4. Types of AUs, translation states and translation phases*

Just like KBs and PUs, also HOF states are made up of AU sequences. Across the 521 translation sessions, there are substantially more AUs in Flow states (F: 70%) than in Hesitation (H: 24%) and Orientation (O: 6%). While almost all types of AUs occur in all HOF states, the proportion and duration vary significantly.

Figure 3 shows bar plots for the proportion (left) and boxplots for the durations (right) of the six types of AUs and the three HOF states. As can be seen in the left plot, Type-8 AUs are dominant in all HOF states, almost 40% of Hesitation, more than 30% of Orientation and 20% of Flow states are Type-8 AUs. The boxplots on the right in Figure 3 show that these AUs have approximately similar durations. Second to Type-8 AUs are Type-4 AUs for Hesitation and Type-1 AUs for Orientation states, while almost 30% of the AUs in Flow states are Type-2 AUs. The duration of Type-1 AUs is by far the longest in orientation states (right box plot). As discussed above, Type-5 AUs are least frequent; they also have the shortest durations in all HOF states (right plot).

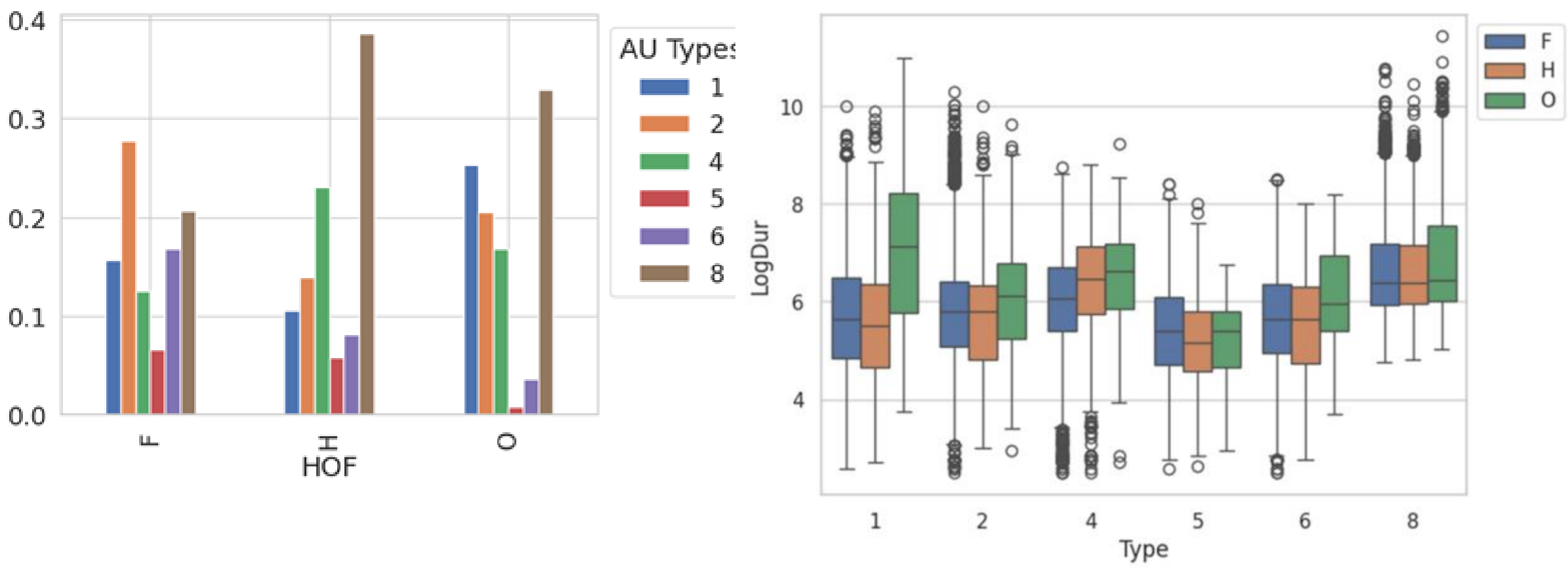

Figure 3: Proportion of AUs across the three HOF states (left) and their log Duration (right)

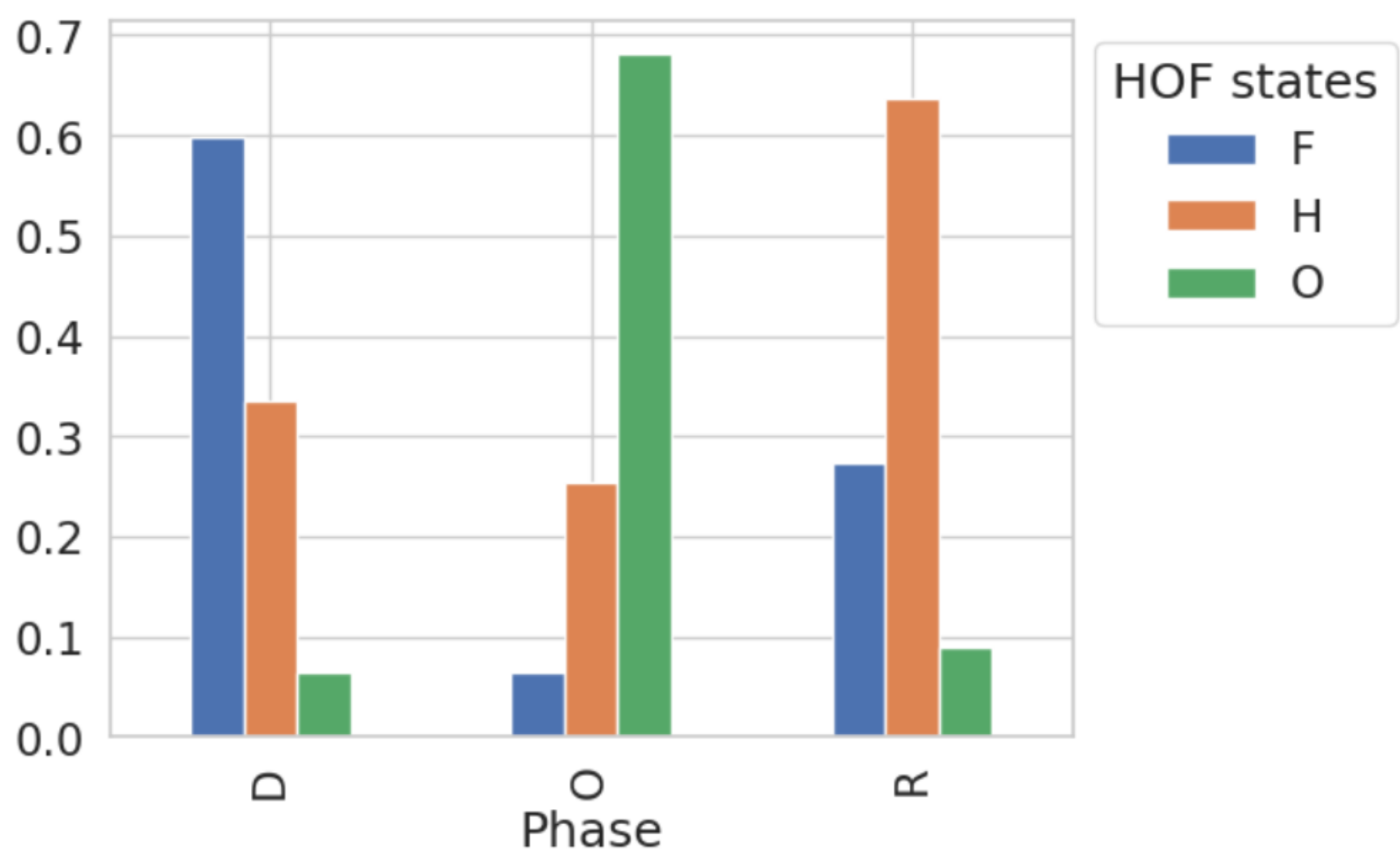

Figure 4: Distribution of HOF states across translation phases

The distribution of HOF states across the three translation phases is shown in Figure 4. The figure shows that each of the three translation phases is dominated by a different HOF state: Over 60% of the states in the orientation phase are Orientations states (O), the drafting phase is dominated by Flow states (F) while states of Hesitations (H) are prevalent in the revision phase. However, there is no clear-cut separation between states and phases, as each of the three translation phases encompasses all three HOF states, albeit to a different extent.

### *4.5. Classifying translator styles*

Mizowaki et al. (2024) have proposed a method to classify personal translation styles (i.e. translator styles) using an unsupervised constrained K-means approach. They use a set of 552 translation sessions from 275 different translators – which largely overlaps with the sessions used in this study – to determine five different types of translator styles. Each translator was automatically assigned a style label that exhibits a particular set of behavioural features. For example, a translator would exhibit a less reflective and more locally oriented translation approach (Dragsted & Carl 2013) if typing activities have frequent interruptions, few insertions and short AU durations. Other translators might show a globally oriented translation style with longer PUBs, fewer deletions and many insertions, suggesting a more deliberate and carefully planned approach. Mizowaki et al. (2024) determine the following five translator styles:

- Style 0: Few insertions, short AU duration, and short KBIs, indicating a rapid, less reflective processing style
- Style 1: Many insertions and long PUBs, indicating a more deliberate, carefully planned style
- Style 2: Very few deletions, suggesting a confident translator, with minimal revisions
- Style 3: Most of the measures are near the overall average, resulting in a balanced style
- Style 4: Many deletions, indicating an extensive revision and more cautious re-evaluation

We have re-used this method to automatically assign translator style labels to our 521 translation sessions. The distribution of translation sessions across the five translator style labels is shown in Figure 5 (left). Most sessions are assigned style label 4, while only few sessions fall under translator style label 0. However, we cannot be certain that the description of our new classification coincides with the one conducted by Mizowaki et al. (2024) since fewer translation sessions were used and the set of features in the BTSS does not exactly coincide with those features used in the Mizowaki et al. (2024) study.

The boxplot on the right side in Figure 5 shows a classification of 60 English-to-Spanish translation sessions produced by 32 different translators[11] according to three translator style labels. The boxplots show the relations between translator styles and LogPUBs values. The average (Log)PUBs seem consistent with the translator style description of Mizowaki et al. (2024) (see above). Individual translator styles depend largely on editing and pausing behaviour (such as KBIs and PUBs), and translators are generally consistent in their pausing behaviour across different translation sessions (Jakobsen 2002; Mizowaki et al. 2024; Carl 2024, 2025). The analysis suggests that translator styles cut across languages and that different translators working in the same language direction (here: English-to-Spanish ) may realize different translator styles.

[11]These sessions are part of BML12 (https://sites.google.com/site/centretranslationinnovation/tpr-db/public-studies,, accessed 12. Sept. 2025). They are also contained in the total of the 521 sessions and in the Mizowaki et al. (2024) study.

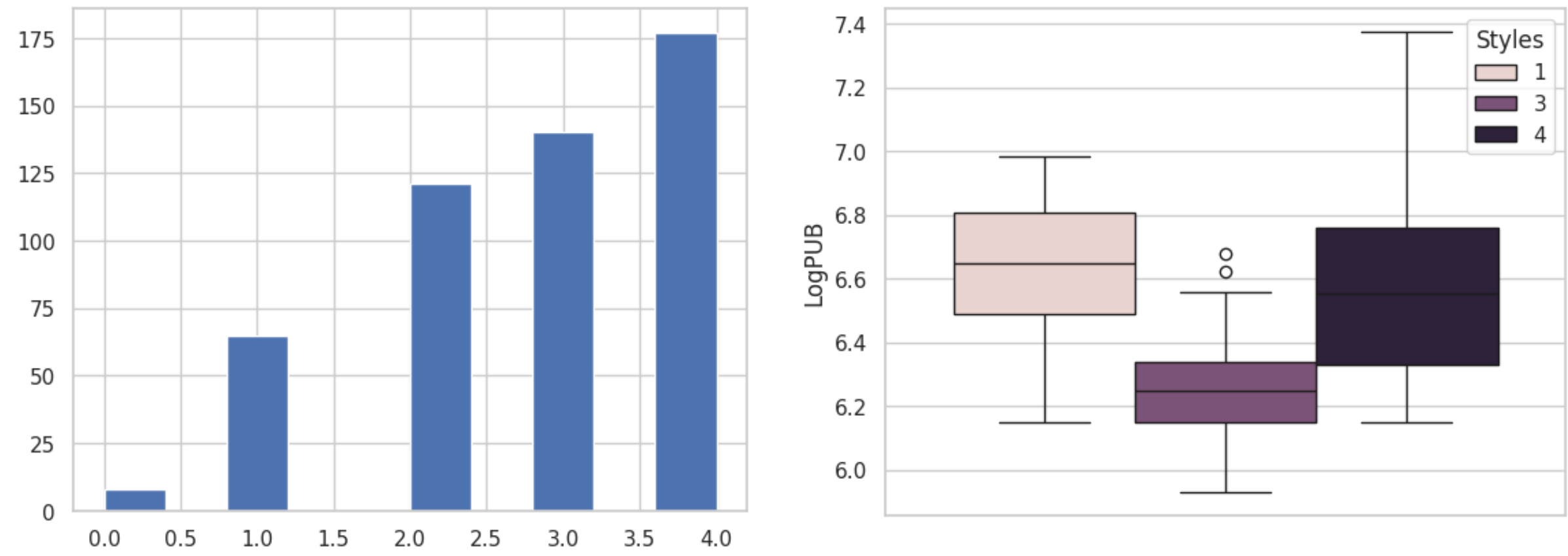

Figure 5: Distribution of translator styles across 521 translation sessions (left) and the classification of 60 Spanish translation sessions and their relation to log PUBs (right).

### *4.6. Relating gaze patterns*

Translator styles, translation phases and HOF states are characterized by typing and pausing behaviours, but the shape of gaze patterns may be equally important as they reveal the translator's visual attention on the ST or the TT. AUs of Types 1 and 5 are characterized by gazing patterns on the ST while gaze patterns on the TT result in AUs of Types 2 or 6. As proposed in Section3.2., gaze patterns can be characterized by their proportion of linear reading (normal, continuous reading behaviour), re-fixations (eyes remains in nearly the same position or move backward), or scattered fixations (no clear reading pattern observed). We disregard thereby which words were fixated but characterize the shape of the gaze patterns and their duration, either as their proportion with respect to the total AU duration (as in Table 5) or in terms of their absolute (log) duration.

Gaze patterns, just like the other processing units of the BTSS layers (AUs, KBs, PUs etc.), are behavioural templates that can be instantiated in concrete translation situations. Thus, the BTSS constitutes a hierarchy of templates that allows us to elaborate, assess, and reproduce a joint probability model of the embedded translation processes. This joint (generative) probability model can subsequently be factorized into specific relations of the translation process in multiple ways.

For instance, the way we introduced AUs of Types 4, 5, or 6 in the BTSS specify how many keystrokes were pressed (and possibly their IKI structure) but they do not specify which keystrokes were precisely produced. Similarly, a gaze path, as we define it here, specifies a pattern of fixation and the number/total duration of the fixations, but it does not provide information about the words that were actually fixated, since these will change from one pattern to the next. While the location of individual fixations on the screen—and thus the gaze-to-word mapping—may not always be very precise in the recordings of translation sessions[12], the relative distances of successive fixation and their durations are reliable parameters of the gaze path templates. Gaze patterns provide us with a level of generalization that allows us to contrast, for instance, the gaze behaviour of an individual translator, or a single translation session or even a small part of it against other sessions.

[12]This may be due to lack or loss of calibration precision (possibly caused by substantial changes in head position as translators switch between different windows), significant vision impairment, the font size may be too small, the line spacing too narrow, and/or the mapping of X/Y fixation points on the screen on the word(s) that where supposedly looked at may be erroneous, among other reasons (cf. Hvelplund 2016).

Figure 6 provides an example that contrasts differences and similarities of the gaze patterns for a particular translation session (BML12/P08_T5, see also Figure 1) against the gaze patterns of the entire 521 sessions. While the overall duration of AUs of translation session P08_T5 resembles the total population distribution (top left), the individual gaze patterns are shorter as compared to the average AUs in our data. This, we suspect, may be one of the features characteristics for translator style 4; it exemplifies how a translation feature of a particular translator deviates from those of the whole population.

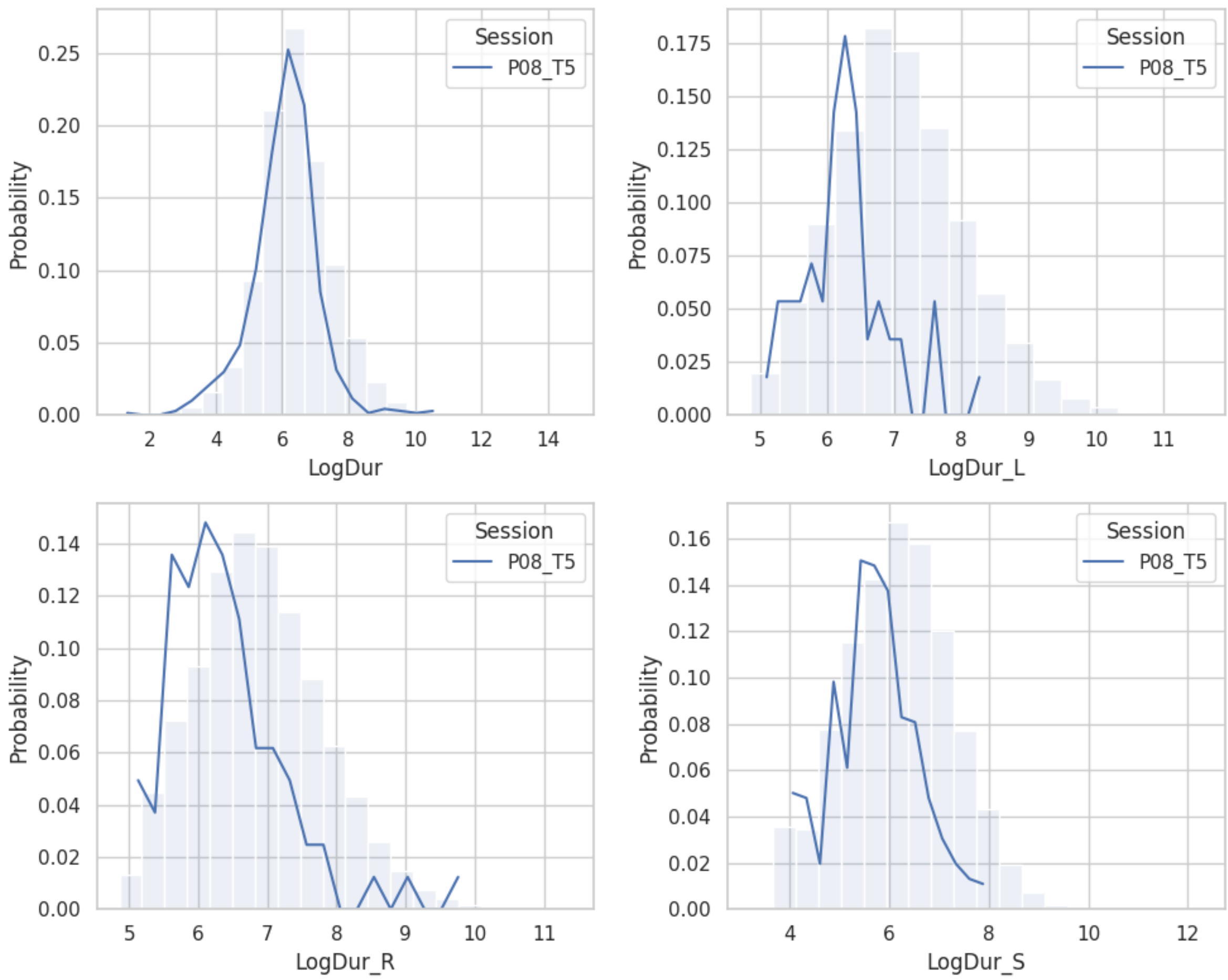

Figure 6: Distribution of gaze patterns for one translation session (BML12/P08_T5) on the background of the lognormal distribution of the entire dataset. The four graphs account for AUs for which the respective durations, Dur, Dur_L, Dur_R, and Dur_S > 1, taking out all AUs that do not contain a gaze patten of this type. While overall the duration of AUs seems to coincide with the overall distribution (top left) the duration of the specific gaze patterns seems on average shorter for P08_T5 than the average.

## 5. Discussion and outlook

Numerous models of the translating mind have been proposed, some of which suggest that several mental processes complement each other during translation production, and many studies have been conducted that support aspects of these models. However, at this stage, the challenge is to develop statistical models that help bridge the gap between data and theory. We aim at addressing this challenge by elaborating a multi-dimensional behavioural translation style space (BTSS) which is designed to subsume all possible behavioural translation patterns[13], and we populate this BTSS with statistical values extracted from empirical translation process data.

The BTSS consists of multi-layered behavioural translation templates that can be serialized and instantiated in concrete translation session. With this multivariate BTSS, we can, on the one hand, define behavioural prototypes and assess behavioural variations across entire populations. On the other hand, we can investigate how behavioural patterns of individual translators deviate from or overlap with those of other translators, how their behaviour relates to (subsets of) population(s), and we can assess possible reasons for these observations. The BTSS allows us to determine specific translation styles by quantifying parameters of how translators traverse the BTSS. While these trajectories instantiate specific translation styles, we can also assess and compare the variation exhibited by individual translators.

Our next step is to implement/refine a computational translation agent (Carl 2024; Mizowaki et al. 2025) that traverses the BTSS on pre-defined trajectories. The agent builds on the free energy principle (FEP, Friston 2009, 2010), predictive processing (PP, Seth 2021; Clark 2023; Hohwy 2016), and active inference (AIF, Friston et al. 2017; Parr et al. 2022). This agent will be suited to address and simulate affective, behavioural/automatized, and/or conscious translation processes.

### *5.1. Translator styles and translation styles*

Several studies have shown that translators develop personal translation styles (which we dub translator styles) characterized by consistent individual behavioural patterns, including typing behaviour, that exhibit affective states, cognitive effort, or decision-making, etc. However, personal translation styles (i.e. translator styles) can be impacted by external parameters, such as familiarity of the text, text difficulty or translation brief. Translators may change their style of translation in line with various textual and contextual parameters, including text register, expected translation quality (i.e. the translation brief), translation directionality (L1 or L2 translation), depending on the translators' expertise, their affective attitude, etc. The realization of a particular translation style thus depends on various parameters that may interact in a non-linear fashion at any moment. Miljanovic et al. (2025), for instance, investigate the relation between individual translator styles, translation directionality (translation into or out of the first language), text register (challenging and less challenging texts), and editing procedures (type of revisions). Their findings suggest a strong interaction between translation direction and register and a large variation between participants. The BTSS may provide a suited framework to explore the scope of this variation on the background of a large corpus and to assess/quantify those (in)variances of translator styles.

[13]Here we only address translation tasks that consists of keystroke and gaze patterns. Future versions of the BTSS might include additional dimensions or processing layers, including for instance, search behaviour (e.g. internet searches, dictionary lookups, etc.), social interaction, communication with peers or customers, etc.

However, given the combinatorial complexity of these interactions and variations, an exhaustive experimental assessment of all possible translation styles and their relation to individual translator preferences seems to be out of reach. However, defining an exhaustive space of continuous features that characterize translation styles might make it possible to specify different translator styles by tracing how BTSS parameter configurations change under varying conditions. Only a subset of the possible interactions may need to be experimentally investigated, thereby providing a limited set of experimentally established points within the BTSS while possible trajectories through the BTSS could then be approximated, given appropriate parameter settings.

### *5.2. Translation agent*

Each of the innumerable instantiations of the BTSS provides a snapshot of possible translation behaviours for any one moment in time. In order to explore that space in a systematic fashion, a computational translation agent would traverse the BTSS and simulate expected translation behaviours for a given piece of text as a specific trajectory. That is, the computational agent would be given a source text with a translation and simulate translation behaviours (i.e. a translation style) that fits a pre-defined configuration, e.g. novice translation, informal translation, etc. The computational agent would thus simulate a detailed account of the (cognitive) effort that different translators would exert when translating the text provided. The agent may also trace or pursue novel translation styles for a set of provided translation settings simulating different levels of expertise, text difficulty, translator preferences, etc. As an agent is, by definition, in constant interaction with its environment, variations and new trajectories of the BTSS may be expected. An appropriate framework will thus be necessary that would evaluate the produced behavioural data by, for instance, comparing it with a gold data set.

We propose the computational agent to be implemented as a multi-layered POMDP (partially observable Markov decision process) using the PyMDP.[14] An available prototype implementation of the agent comprises three hierarchical layers—affective (A), behavioural (B) and cognitive (C) —each operating on a distinct time scale and jointly govern translation behaviour. The A-layer captures the translator's emotional states as specified through the HOF taxonomy. The C-layer accounts for deliberation and planning that result in PUBs. The B-layer controls concrete translation actions (e.g. typing or eye movements) and tracks progress via Alignment Groups that map ST tokens to their corresponding TT tokens.

By framing the agent as a POMDP, the system receives observable cues while estimating the translator's unobservable affective and cognitive states. The PyMDP API offers multiple observation modalities and hidden state factors that may be suited to address multiple dimensions of the BTSS. The relationship between hidden state factors and observation modalities is modelled through an observation likelihood. This observation likelihood specifies the probability of observing a particular input, for instance four words in linear reading, given the hidden state. The transition between hidden states is conditioned on actions, such as typing. On the one hand, actions indirectly influence (future) observation modalities by altering hidden state factors. Hidden state factors, thus, determine how observations are generated. On the other hand, actions are chosen to minimize the (expected) free energy for each translation style parameter independently.

[14]A Python library for modelling Markov Decision Processes is:
https://github.com/infer-actively/pymdp/blob/master/docs/notebooks/using_the_agent_class.ipynb

At each time step, the agent can update its internal states based on observations in a Bayesian fashion and select the next action (e.g. continue typing, pause for a more re-reading of the ST, or revise recently typed TT). This choice is guided by active inference, which aims to minimize future ambiguity and cost (i.e. expected free energy). Through repeated cycles of observation, state estimation, and action selection over embedded layers of the hierarchical architecture, the agent recreates the temporal dynamics of the translation process. Furthermore, nested active inference loops across the three ABC layers enable the model to capture differences in temporal granularity among affective, cognitive, and behavioural processes.

### *5.3. Bottom-up vs. top-down processing*

PP posits two complementary processing directions: 1) bottom-up processes, in which observations may generate prediction errors when they mismatch top-down predictions. 2) top-down processes in which higher-level cognitive states, such as expectations, beliefs, or priors, influence how sensory inputs are interpreted. The resolution of these processes is mediated by a precision parameter, which modulates the balance between sensory-driven (bottom-up) and model-driven (top-down) influences. This bidirectional interplay is essential for adaptive and efficient agent behaviour. Actions further integrate these processes by enabling active exploration, reducing uncertainty, and optimizing perception and decision-making.

The affective and cognitive layers of the agent play a crucial role in altering precision parameters in different ways. Affective states (e.g. anxiety, motivation, surprise, curiosity) may increase precision for sensory inputs, heightening vigilance (bottom-up emphasis). In contrast, cognitive processes (e.g. reasoning, planning) would refine predictions and update hidden state representations (top-down influence), including predictions about grammatical structures sentence formulation (see Lehka-Paul 2020, Section 2.5.). The two layers may engage in an emotion-cognition interaction in which affective states regulate precision allocation, impacting cognitive strategies and behavioural outputs. Precision parameters may then dynamically weight the influence of bottom-up vs. top-down signals, adjusted by affective states while cognitive processes shape predictions, and behaviour enacting corrections.

### *5.4. Outlook*

At present, numerous assumptions of the agent's functioning remain provisional, as they have not yet undergone direct optimization using empirical data. In future, we intend to base the observation likelihood parameters and transition probabilities on distributions derived from BTSS. Another technical challenge pertains to how the model should control the duration of each state. For instance, it remains unclear how long a Flow state should persist or when C-layer reflective activities would transition back to a Flow mode. What information should precisely be transmitted between the different layers and at what point in time. Although the BTSS specifies individual differences in the translators' working styles, our current agent implementation does not yet provide a fully developed set of probabilistic parameters to capture these variations. Further theoretical development and data analysis will be required to refine how state transition probabilities and likelihoods can account for diverse translation styles within the model. By asking what triggers transitions between those different mental processing layers, how do translators plan ahead, and why do automatized routines shift to reflective thought or revision, the suggested computational agent integrates insights from cognitive science and bilingualism (Van Gompel 2013; Kaan & Grüter 2021) into the dynamics of cognitive control, memory, and planning to evaluate and enhance its theoretical foundation.

We believe that the suggested topology of the BTSS in conjunction with a PP/AIF-based computational translation agent has the potential to generate novel insights into the interactions between affective, reflective, and routinized translation processes in expert and novice translators, and may establish a game changing paradigm shift for investigating the effects of feeling and thinking of the translating mind. A simulation approach in cognitive translation and interpretation studies (CTIS) might thereby bridge the gap with broader enquiries in cognitive science and bilingualism, advancing our understanding of the interplay between intuitive and deliberative mental processes.

## Data Availability Statement

The data used in this article are freely available and can be downloaded from the CRITT website https://sites.google.com/site/centretranslationinnovation/tpr-db/ The Python scripts that produces the BTSS can be downloaded from GitHub https://github.com/Critt-Kent/Behavioral-Translation-Style-Space (accessed 2025-12-09).

## References

Ajzen, Icek. 1991. The theory of planned behavior. *Organizational Behavior and Human Decision Processes* 50(2): 179–211.

Alves, Fabio & Vale, Daniel C. 2011. On drafting and revision in translation. A corpus linguistics-oriented analysis of translation process data. In: Hansen-Schirra, Silvia & Neumann, Stella & Čulo, Oliver (eds.), *Annotation, exploitation and evaluation of parallel corpora*. Berlin: Language Science Press. 89–110.

Alves, Fabio & Albir, Amparo Hurtado. 2025. *Translation as a cognitive activity. Theories, models and methods for empirical research*. Routledge.

Angelone, Erik. 2010. Uncertainty, uncertainty management and metacognitive problem solving in the translation task. In: Angelone, Erik & Shreve, Gregory M. (eds.), *Translation and cognition*. Amsterdam: John Benjamins Publishing. doi: 10.1075/ata.xv.03ang.

Bernardini, Silvia. 2001. Think-aloud protocols in translation research. Achievements, limits, future prospects. *Target* 13(2): 241–63.

Briggs Myers, I. 1962. *The Myers-Briggs type indicator: Manual*. Palo Alto: Consulting Psychologists Press.

Carl, Michael & Kay, Martin. 2011. Gazing and Typing Activities during Translation : A Comparative Study of Translation Units of Professional and Student Translators. In: Meta, Vol. 56, No. 4, 2011, p. 952-975.Carl, Michael. 2021. Micro units and the first translational response universal. In Carl, Michael (ed.), *Explorations in empirical translation process research*. Cham: Springer. 233–57.

Carl, Michael & Wei, Yuxiang & Lu, Sheng & Zou, Longhui & Mizowaki, Takanori & Yamada, Masaru. 2024. Hesitation, orientation, and flow. A taxonomy for deep temporal translation architectures. *Ampersand* 12: 100164.

Carl, Michael & Dragsted, Barbara & Lykke Jakobsen, Arnt. 2011. A Taxonomy of Human Translation Styles. *Translation Journal* 16(2).

Carl, Michael & Schaeffer, Moritz & Bangalore, Srinivas. 2016. The CRITT Translation Process Research Database. In Carl, Michael & Bangalore, Srinivas & Schaeffer, Moritz (eds.), *New directions in empirical translation process research.* Cham: Springer. 13–54.

Carl, Michael. 2023. Models of the translation process and the free energy principle. *Entropy* 25(6): 928. doi: 10.3390/e25060928.

Carl, Michael. 2024. An active inference agent for modeling human translation processes. *Entropy* 26(8): 616.

Carl, Michael & Schaeffer, Moritz. 2017. Sketch of a noisy channel model for the translation process. Empirical modelling of translation and interpreting. In Hansen-Schirra, Silvia, Czulo, Oliver, Hofmann, Sascha (eds.), *Translation and multilingual natural language processing* (No. 7). Berlin: Language Science Press. 71–116.

Carl, Michael. (2025) Temporal dynamics of emotion and cognition in human translation. Integrating the task segment framework and the HOF taxonomy. *Digital Studies in Language and Literature* doi: 10.1515/dsll-2025-0002.

Clark, Andy. 2023. The experience machine: How our minds predict and shape reality. New York: Pantheon Books.

Clifton Jr, Charles & Staub, Adrian & Rayner, Keith. 2007. Eye movements in reading words and sentences. In Gaskell, M. Gareth (ed.), *The Oxford handbook of psycholinguistics*. Oxford University Press. 341–55.

Dijkstra, Ton, & Van Heuven, Walter J. B. 2002. The architecture of the bilingual word recognition system. From identification to decision. *Bilingualism. Language and Cognition* 5(3): 175–97. doi: 10.1017/s1366728902003012.

Dragsted, Barbara. 2005. Segmentation in translation. Differences across levels of expertise and difficulty. *Target* 17(1): 49–70.

Dragsted, Barbara & Carl, Michael. 2013. Towards a classification of translation styles based on eye-tracking and keylogging data. *Journal of the Writing Research* 5(1): 133–58.

Englund-Dimitrova, Birgitta. 2005. *Expertise and explicitation in the translation process*. Amsterdam: John Benjamin Publishing Company.

Evans, Jonathan St. B. T. & Stanovich, Keith E. 2013. *Dual process theories of cognition. Advancing the debate*. Perspectives on Psychological Science 8(3): 223–41.

Friston, Karl. 2009. The free-energy principle. A rough guide to the brain? *Trends in Cognitive Sciences* 13(7): 293–301.

Friston, Karl J. & Lin, Marco & Frith, Christopher D. & Pezzulo, Giovanni & Hobson, J. Allan & Ondobaka, Sasha. 2017. Active inference, curiosity and insight. *Neural Computation* 29(10): 2633–83. doi: 10.1162/neco_a_00999.

Friston, Karl. 2010. The free-energy principle. A unified brain theory? *Nature Reviews Neuroscience* 11: 127–38. doi: 10.1038/nrn2787.

Gutt, Ernst-August. 2000. Translation and relevance. Cognition and context. Manchester: St. Jerome.

Gutt, Ernst-August. 2005. On the significance of the cognitive core of translation. *The Translator* 11(1): 25–49.

Hansen, Gyde. 2013. Many tracks lead to the goal. A long-term study on individual translation styles. In Way, Catherine & Vandepitte, Sonia & Meylaerts, Reine & Bartlomiejczyk, Magdalena (eds.), *Tracks and treks in translation studies*. Amsterdam: John Benjamins Publishing Company. 49–62.

Heins, Connor & Millidge, Beren & Demekas, Daphne & Klein, Brennan & Friston, Karl & Couzin, Iain & Tschantze, Alexander. 2022. Pymdp: A Python library for active inference in discrete state spaces. *Journal of Open Source Software* 7(73): 4098. doi: 10.21105/joss.04098.

Hohwy, Jakob. 2016. The self-evidencing brain. *Nous* 50(2): 259–85.

Hubscher-Davidson, Séverine. 2017. *Translation and emotion. A psychological perspective*. New York: Routledge.

Hvelplund, Kristian T. 2016. Cognitive efficiency in translation. In Muñoz, Ricardo Martín (ed.), *Reembedding translation process research*. Amsterdam: John Benjamins Publishing Company. pp. 149–170

Jakobsen, Arnt L. 2002. Translation drafting by professional translators and translation students. In Hansen, Gyde (ed.), *Empirical translation studies. Process and product* (Copenhagen Studies in Language 27). Copenhagen: Samfundslitteratur. 191–204.

Jakobsen, Arnt L. 2011. Tracking translators' keystrokes and eye movements with Translog. In Alvstad, Cecilia & Hild, Adelina & Tiselius, Elisabet (eds.), *Methods and strategies of process research. Integrative approaches in translation studies*. Amsterdam: John Benjamins Publishing Company. 37–55.

Jakobsen, Arnt L. 1999. Logging target text production with Translog. *Copenhagen Studies in Language* 24: 9–20.

Jung, Carl Gustav. [1921] 1971. *Psychological types. Collected works of Carl Gustav Jung* (Volume 6). Princeton, NJ: Princeton University Press.

Kaan, Edith & Grüter, Theres. 2021. *Prediction in second language processing and learning. Bilingual processing and acquisition.* Amsterdam: John Benjamins Publishing Company.

Kahneman, Daniel. 2011. *Thinking, fast and slow*. Macmillan.

Lacruz, Isabel & Shreve, Gregory M. 2014. Pauses and cognitive effort in post-editing. In O'Brien, Sharon & Winther Balling, Laura & Carl, Michael & Simard, Michel & Specia, Lucia (eds.), *Post-editing of machine translation. Processes and applications*. Cambridge Scholars Publishing.

Lehka-Paul, Olha. 2020. Behavioural indicators of translators' decisional styles in a translation task. A longitudinal study. *Yearbook of the Poznań Linguistic Meeting* 6: 81–111. doi: 10.2478/yplm-2020-0007.

Malmkjaer, Kristen. 1998. Unit of Translation. In Baker, Mona & Malmkjaer, Kristen (eds.), *Routledge encyclopedia of translation studies*. London: Routledge. 286–88.

Miljanovic, Zoë & Alves, Fabio & Brost, Celina & Neumann, Stella. 2025. Directionality in translation. Throwing new light on an old question. *SKASE Journal of Translation and Interpretation* (this issue.)

Mirlohi, Mehdi & Egbert, Joy & Ghonsooly, Behzad. 2011. Flow in translation. Exploring optimal experience for translation trainees. *Target* 23(2): 251–71. doi: 10.1075/target.23.2.06mir.

Mizowaki, Takanori & Yamada, Masaru & Carl, Michael & Wei, Yuxiang. 2024. Modeling and simulation of the translation process using hidden Markov models. (Paper presented at the *14th International and Interdisciplinary Conference on Applied Linguistics and Professional Practice (ALAPP)*, Hong Kong, 17–19 December 2024.)

Mizowaki, Takanori & Yamada, Masaru & Carl, Michael & Wei, Yuxiang & Raj, Aishvarya & Bandaru, Devi Sri & Ren, Xinyue. 2025. Building and evaluating an active-inference model of the translation process. A simulation for quantifying translators' 'surprise'. (Paper presented at the 26th Japan Association for Interpreting and Translation Studies (JAITS) Annual Conference, Institute of Science Tokyo, Tokyo, 6–7 September 2025.)

Mossop, B. 2007. Empirical studies of revision. What we know and need to know. *The Journal of Specialised Translation* 8: 5–20.

Muñoz, Ricardo & Apfelthaler, Matthias. 2022. A task segment framework to study keylogged translation processes. *Translation & Interpreting* 14(2): 8–31.

Rosa, Rusdi Noor & Sinar, T. Silvana & Ibrahim-Bell, Zubaidah & Setia, Eddy. 2018. Pauses by student and professional translators in translation process. International Journal of Comparative Literature & Translation 6(1). https://journals.aiac.org.au/index.php/IJCLTS/article/view/4107 (Accessed 2025-12-09.)

Parr, Thomas & Pezzulo, Giovanni & Friston, Karl J. 2022. *Active inference: The free energy principle in mind, brain, and behavior*. MIT Press.

Rayner, Keith. 1998. Eye movements in reading and information processing. 20 years of research. *Psychological Bulletin* 124(3): 372–422.

Robinson, Douglas. 2023. *Questions for Translation Studies*. London: Routledge.

Rojo López, Ana & Caldwell-Harris, Catherine. 2023. Emotions in cognitive translation and interpreting studies. In Saldanha, Gabriela & O'Brien, Sharon (eds.), *The Routledge handbook of translation, interpreting and bilingualism*. Manchester: St Jerome. doi: 10.4324/9781003109020-18

Schaeffer, Moritz & Tardel, Anke & Hofmann, Sascha & Hansen-Schirra, Silvia. 2019. Cognitive effort and efficiency in translation revision. In Huertas-Barros, Elsa & Vandepitte, Sonia & Iglesias-Fernández, Emilia (eds.), *Quality assurance and assessment practices in translation and interpreting*. Hershey: IGI Global. 226–43.

Schaeffer, Moritz & Carl, Michael. 2013. Shared representations and the translation process. A recursive model. *Translation and Interpreting Studies* 8(2): 169–90.

Schotter, Elizabeth R. & Angele, Bernhard & Rayner, Keith. 2012. Parafoveal processing in reading. *Attention, Perception, & Psychophysics* 74(1): 5–35.

Sperber, Dan & Wilson, Deirdre. [1986] 1995. *Relevance. Communication and cognition* (2nd edition). Oxford: Blackwell. (Page references to the second edition from 1995.)

Thunes, Martha. 2017. The concept of 'translation unit' revisited. *Bergen Language and Linguistics Studies* 8. doi: 10.15845/bells.v8i1.1331.

Tirkkonen-Condit, Sonja. 2005. The monitor model revisited. Evidence from process research. *Meta* 50(2): 405–14. doi: 10.7202/010990ar.

Van Gompel, Roger P. G. 2013. *Sentence processing. An introduction.* New York: Psychology Press.

*Michael Carl*
*Modern and Classical Language Studies*
*Kent State University, USA*
*e-mail: mcarl6@kent.edu*

*Takanori Mizowaki*
*Graduate School of Intercultural Communication*
*Rikkyo University, Japan*
*e-mail: 22wv006d@rikkyo.ac.jp*

*Devi Sri Bandaru*
*College of Arts and Sciences*
*Kent State University, USA*
*e-mail: dbandaru@kent.edu*

*Xinyue REN*
*Department of Linguistics and Translation*
*City University of Hong Kong, HKSAR*
*e-mail: xinyueren3-c@my.cityu.edu.hk*

*Aishvarya Raj*
*Independent researcher*
*London, United Kingdom*
*e-mail: aishvaryaraj@live.co.uk*

*Masaru Yamada*
*College/Graduate School of Intercultural Communication*
*Rikkyo University, Japan*
*e-mail: masaru.yamada@rikkyo.ac.jp*